\documentclass[a4paper, conference]{IEEEtran}

\IEEEoverridecommandlockouts

\usepackage{float}
\usepackage{wrapfig}
\usepackage{float}
\usepackage{cite}
\usepackage{amsmath,amssymb,amsfonts}
\usepackage{algorithm}
\usepackage{algorithmic}
\usepackage{graphicx}
\usepackage{textcomp}
\usepackage{xcolor}
\def\BibTeX{{\rm B\kern-.05em{\sc i\kern-.025em b}\kern-.08em
    T\kern-.1667em\lower.7ex\hbox{E}\kern-.125emX}}
\begin{document}

\title{Enhancing Object Detection Robustness:
Detecting and Restoring Confidence in the Presence of Adversarial Patch Attacks
}

\author{\IEEEauthorblockN{1\textsuperscript{st} Roie Kazoom}
\IEEEauthorblockA{\textit{Electrical and Computers Engineering} \\
\textit{Ben Gurion University of The Negev}\\
Beer Sheva, Israel \\
roieka@post.bgu.ac.il}
\and
\IEEEauthorblockN{2\textsuperscript{rd} Raz Birman}
\IEEEauthorblockA{\textit{Electrical and Computers Engineering} \\
\textit{Ben Gurion University of The Negev}\\
Beer Sheva, Israel \\
birmanr@post.bgu.ac.il}
\and
\IEEEauthorblockN{3\textsuperscript{nd} Ofer Hadar}
\IEEEauthorblockA{\textit{Electrical and Computers Engineering} \\
\textit{Ben Gurion University of The Negev}\\
Beer Sheva, Israel \\
hadar@bgu.ac.il}
}

\maketitle

\begin{abstract}

The widespread adoption of computer vision systems has underscored their susceptibility to adversarial attacks, particularly adversarial patch attacks on object detectors. This study evaluates defense mechanisms for the YOLOv5 model against such attacks. Optimized adversarial patches were generated and placed in sensitive image regions, by applying EigenCAM and grid search to determine optimal placement. We tested several defenses, including Segment and Complete (SAC), Inpainting, and Latent Diffusion Models. Our pipeline comprises three main stages: patch application, object detection, and defense analysis. Results indicate that adversarial patches reduce average detection confidence by 22.06\%. Defenses restored confidence levels by 3.45\% (SAC), 5.05\% (Inpainting), and significantly improved them by 26.61\%, which even exceeds the original accuracy levels, when using the Latent Diffusion Model, highlighting its superior effectiveness in mitigating the effects of adversarial patches.

\end{abstract}


\section{Introduction}
Computer vision systems have become increasingly pervasive in domains such as autonomous vehicles, intelligent security cameras, and industrial automation. Central to these systems are object detectors, which play a crucial role in accurately identifying and tracking objects. However, the reliance on these technologies has exposed them to significant security risks, particularly in the form of adversarial attacks. Among these, adversarial patch attacks are a notable threat, where small, maliciously designed patches are added to images to deceive object detection models, potentially leading to incorrect or missed detections and severe consequences in real-world applications \cite{hwang2023gan}.
Moreover, recent work has shown that retrieval-augmented generation can detect adversarial manipulations without additional training \cite{kazoom2025dontlag}, and that meta-classification models can achieve strong performance even on small datasets \cite{kazoom2022meta}.

As adversarial patch attacks grow more sophisticated, developing robust defense mechanisms has become critical to ensure the reliability and security of object detection systems, especially in high-stakes environments. This study addresses the challenge of defending against adversarial patch attacks on the YOLOv5 model, a widely used, state-of-the-art object detector.

The primary contribution of this work is a comprehensive pipeline to systematically evaluate different defense strategies against adversarial patches. This pipeline focuses on patch mask defenses, allowing for detailed analysis of their effectiveness. The defense methods include the Segment and Complete (SAC) algorithm, utilizing a U-Net architecture to identify and remove adversarial patches; an inpainting algorithm that replaces the patch area by interpolating surrounding pixel values; and a stable diffusion model that reconstructs high-resolution images from latent representations within a variational autoencoder (VAE), employing U-Net with convolutional layers, skip connections, and attention mechanisms. These methods were selected for their potential to restore detection performance compromised by adversarial patches.

This paper contributes to both attack and defense methodologies. For the attack, we apply a post-optimization step after creating the patch, enhancing its effectiveness compared to classical methods that optimize only a single function. On the defense side, our approach advances beyond traditional techniques by combining classical segmentation for patch detection and image processing for restoration with a latent diffusion model to restore occluded parts of the image. The latent diffusion model improves computational efficiency, essential for real-time adversarial patch removal. While classical methods approximate the original confidence rate of an object, our method improves upon it, as the diffusion model often reconstructs object features better than the original image.

This research contributes to safeguarding object detection systems against adversarial attacks by providing a validated pipeline for testing defense mechanisms. The findings highlight the effectiveness of robust defenses in enhancing the resilience of computer vision systems and suggest promising directions for future work aimed at further improving security and reliability across diverse applications.

\section{Literature Review}
Several studies have proposed various methods for generating adversarial patches, each contributing to the growing body of research in this area. This section reviews the state-of-the-art techniques in both generating adversarial attacks and the corresponding defense mechanisms developed to mitigate them.

\subsection{Adversarial Patch Attacks}
Hwang et al. \cite{hwang2023gan} proposed a GAN method for generating adversarial patches to carry out dodging and impersonation attacks on face recognition systems. Liu et al. \cite{liu2024rpa} Introduced three robust physical attacks against Unmanned Aerial Vehicles (UAVs): Hiding Attack, Yaw Attack, and Obstacle Attack. Deng et al. \cite{deng2023rust} developed a natural rust-styled adversarial patch generation method with a small perturbation area for remote sensing images used in military applications. Their patch generation framework is based on style transfer. 

Lapid et al. \cite{lapid2023patch} proposed a black-box, gradient-free method that uses the learned image manifold of a pretrained GAN to generate naturalistic physical adversarial patches to mislead object detectors. Their approach outperformed other tested methods by a significant margin. Lin et al. \cite{lin2023diffusion} introduced a novel naturalistic adversarial patch generation method based on diffusion models, which can stably craft high-quality and naturalistic physical adversarial patches visible to humans, without suffering from serious mode collapse problems. Zhou et al. \cite{zhou2024mvp} presented MVPatch, an approach aimed at improving the transferability, stealthiness, and naturality of adversarial patches using an ensemble attack loss function and a visual similarity measurement algorithm realized by the CSS loss function. These developments build upon the research summarized by Wei et al. \cite{wei2022survey}, highlighting the evolution of adversarial patch attacks over the last decade.

\subsection{Defense Mechanisms}
In response to these adversarial attacks, various defense mechanisms have been proposed to protect object detectors from adversarial patches. Chua et al. \cite{chua2022duet} introduced an uncertainty-based adversarial patch localizer, allowing for post-processing patch avoidance or reconstruction. Their method quantifies prediction uncertainties using the Detection of Uncertainties in the Exceedance of Threshold (DUET) algorithm, providing a framework to ascertain confidence in adversarial patch localization. Liu et al. \cite{liu2022sac} proposed Segment and Complete defense (SAC), a general framework for defending object detectors against patch attacks through the detection and removal of adversarial patches. They trained a patch segmenter and proposed a self-adversarial training algorithm to enhance its robustness. Additionally, they designed a robust shape completion algorithm to guarantee the removal of the entire patch under specific conditions. Taeheon et al. \cite{kim2022ape} introduced a defense method based on "Adversarial Patch - Feature Energy" (APE), exploiting common deep feature characteristics of adversarial patches. Their defense consists of APE-masking and APE-refinement, which can be applied to any adversarial patch. Yang et al. \cite{yang2023ibcd} proposed a two-stage Iterative Black-box Certified Defense method, termed IBCD. The first stage estimates the patch size using a search-based method by evaluating the size relationship between the patch and mask with pixel masking. The second stage calculates accuracy using existing white-box defense methods with the estimated patch size. Kang et al. \cite{kang2024diffender} introduced DIFFender, a novel defense method leveraging a text-guided diffusion model to defend against adversarial patches, comprising patch localization and patch restoration stages. Xue et al. \cite{xue2024dpg} presented a Defense Patch GNN (DPG) to counter adversarial patch attacks. They extracted and downsampled image features, introduced a graph-structured feature subspace for enhanced robustness, and designed an optimization algorithm using Stochastic Gradient Descent. Their model showed superior robustness against existing adversarial patch attacks. Liang et al. \cite{liang2024texture} proposed a defense method based on texture features and local denoising, detecting the location of potential adversarial patches and subsequently denoising the image. Pathak et al. \cite{pathak2024agnostic} proposed a model-agnostic defense against adversarial patch attacks in UAV-based object detection, treating patches as occlusions and removing them without prior exposure during training. Lin et al. \cite{lin2024nutnet} introduced NutNet, an innovative model for detecting adversarial patches, demonstrating high generalization, robustness, and efficiency. Their method effectively defended against both Hiding and Appearing Attacks on six detectors, including YOLOv2-v4, SSD, Faster RCNN, and DETR, in both digital and physical domains with minimal performance loss.

\section{Methodology}
In this study, YOLOv5 was selected as the target model for evaluating various defense techniques against adversarial patch attacks, given its popularity and performance in the field. This choice enables a direct comparison with other works, as YOLOv5 is widely used in adversarial robustness studies. The attack optimization was performed using the methods proposed in our previous work \cite{kazoom2024improving}, which focus on enhancing the success rate and robustness of adversarial patches. Several attacks were successfully generated and applied to the object detection pipeline following the optimized approach. Following further refinement attempts, we utilized the patch placement strategy suggested in the referenced methods, which recommended positioning the patch at the center of the object's bounding box. This strategy consistently resulted in effective attacks, aligning with other findings in the literature \cite{liu2022sac}.

\subsection{Patch Generation}
An adversarial patch is created by defining an objective function $\mathcal{L}$, typically the cross-entropy loss, which is optimized to ensure that the image with the patch is classified as the target class $t$. This function often includes a regularization term to ensure that the perturbation $\delta$ added to the input image $x$ guides the model $f$ towards the target classification, while keeping the perturbation magnitude under control with a parameter $\lambda$. The objective can be represented as:

\begin{equation}
\mathcal{L}(f(x + \delta), t) + \lambda \|\delta\|_p,
\end{equation}

where $\|\delta\|_p$ is the $p$-norm of the perturbation, used to regularize the patch size. Starting with an initial random patch $\delta$, the optimization loop iteratively updates the patch by applying a gradient descent step:

\begin{equation}
\delta \leftarrow \delta - \eta \frac{\partial \mathcal{L}}{\partial \delta},
\end{equation}

where $\eta$ is the learning rate. After each update, the patch values are clipped to ensure they remain within valid pixel ranges. This process is repeated until the objective function converges or a predetermined number of iterations is reached, resulting in an adversarial patch that causes the model to misclassify the object as the target \cite{hwang2023gan}.

\subsection{Patch Optimization Using Grid Search and EigenCAM}

To optimize the placement of adversarial patches in object detection, we employed two key techniques: EigenCAM \cite{muhammad2020eigen} and Grid Search.

\subsubsection{EigenCAM}
EigenCAM \cite{muhammad2020eigen} identifies key regions that contribute most to the model's classification decision by analyzing the feature maps of the last convolutional layer. Let \( I \in \mathbb{R}^{i \times j} \) denote the input image and \( W_L \) represent the weight matrix of the model at layer \( L \). The output from this layer can be projected as:

\begin{equation}
    O_{L=k} = W_{L=k}^T I
\end{equation}

We apply Singular Value Decomposition (SVD) on \( O_{L=k} \), yielding:

\begin{equation}
    O_{L=k} = U \Sigma V^T
\end{equation}

Where \( U \), \( \Sigma \), and \( V \) are the matrices containing the left singular vectors, singular values, and right singular vectors, respectively. The class activation map, \( L_{\text{Eigen-Cam}} \), is then derived by projecting \( O_{L=k} \) onto the first eigenvector:

\begin{equation}
    L_{\text{Eigen-Cam}} = O_{L=k} V_I
\end{equation}

This highlights the regions most relevant to the model’s decision-making process, indicating the optimal areas where adversarial patches can be placed to maximize their impact.

\subsubsection{Grid Search}
Following the identification of key regions using EigenCAM, we applied Grid Search to optimize the exact location of the adversarial patch. Grid Search involves placing the patch at various candidate locations within the identified region and evaluating the model’s response. Let \( C(x, \delta) \) represent the confidence score of the model for the input image \( x \) with patch \( \delta \). For each possible location \( p_i \), the patch is placed at location \( p_i \), and the corresponding model confidence \( C(x, \delta_{p_i}) \) is computed:

\begin{equation}
    \delta_{p_{\text{optimal}}} = \arg \min_{p_i} C(x, \delta_{p_i})
\end{equation}

This process iteratively evaluates all positions within the targeted region and selects the patch location \( p_{\text{optimal}} \) that minimizes the model's confidence score, thereby maximizing the adversarial impact on object detection.

\begin{figure}[H]
    \centering
    \includegraphics[width=0.475\textwidth]{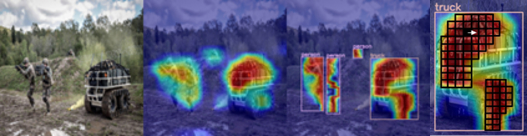}
    \caption{Visualization of the adversarial patch application process: (a) Original image, (b) EigenCAM heatmap highlighting areas of importance, (c) Heatmap with bounding box around the target object, (d) Grid search overlay for optimal patch placement on a specific object.}
    \label{fig:combined}
\end{figure}

By combining EigenCAM's spatial feature analysis with Grid Search, we ensure the patch is both strategically placed and effective at reducing the model's classification confidence \cite{kazoom2024improving}. Figure~\ref{fig:combined} illustrates the step-by-step process of the adversarial patch application, starting from the original image and ending with the optimized patch placement.

\subsection{Adversarial Patch Attack Pipeline}

Algorithm~\ref{alg:patch_attack} outlines the procedure for generating and deploying an adversarial patch aimed at degrading the performance of an object detection model. Let \( f_\theta : \mathcal{X} \rightarrow \mathcal{Y} \) be a pre-trained object detector parameterized by \( \theta \), where \( \mathcal{X} \) is the input image space and \( \mathcal{Y} \) the output label space. The goal is to construct a localized perturbation \( \delta \in \mathbb{R}^{H_p \times W_p \times 3} \), constrained to \( \|\delta\|_\infty \leq \epsilon \), such that when applied to an input image \( x \in \mathcal{X} \), the perturbed image \( x' = x \oplus_{p} \delta \) leads to a degradation in detection performance.

The attack is formalized as the following constrained optimization problem:
\[
\delta^* = \arg\max_{\delta \in \mathcal{S}} \; \mathbb{E}_{x \sim \mathcal{D}} \left[ \mathcal{L}(f_\theta(x \oplus_{p} \delta), y) \right],
\]
where \( \mathcal{S} \) denotes the feasible set (e.g., pixel and norm constraints), \( \mathcal{L} \) is the detection loss, and \( y \) is the ground-truth label associated with \( x \). The operation \( \oplus_{p} \) denotes the application of patch \( \delta \) at position \( p \).

To enhance attack effectiveness, we first localize a set of salient candidate regions \( \{R_1, \ldots, R_k\} \subset \mathcal{X} \) using model interpretability techniques such as EigenCAM. A discrete set of potential patch locations \( \mathcal{P} = \{p_1, \ldots, p_n\} \) within these regions is then evaluated using grid search:
\[
p^* = \arg\max_{p_i \in \mathcal{P}} \; \mathcal{L}(f_\theta(x \oplus_{p_i} \delta), y).
\]

Finally, the patch is refined via iterative gradient-based optimization. At each iteration \( t \), we update:
\[
\delta^{(t+1)} \leftarrow \Pi_{\mathcal{S}} \left( \delta^{(t)} + \eta \cdot \nabla_\delta \mathcal{L}(f_\theta(x \oplus_{p^*} \delta^{(t)}), y) \right),
\]
where \( \eta \) is the learning rate and \( \Pi_{\mathcal{S}} \) is a projection operator ensuring the patch remains within the feasible set. This process continues until convergence or a fixed iteration budget is met, resulting in a robust adversarial patch \( \delta^* \) that consistently degrades the model’s predictions when applied at the optimized location \( p^* \).

\begin{algorithm}[H]
\caption{Adversarial Patch Attack using EigenCAM and Grid Search}
\label{alg:patch_attack}
\begin{algorithmic}[1]
\renewcommand{\algorithmicrequire}{\textbf{Input:}}  
\renewcommand{\algorithmicensure}{\textbf{Output:}}  

\REQUIRE Image $I \in \mathbb{R}^{i \times j}$, class $t$, model $f$, patch size $\delta$, learning rate $\eta$, iterations $N$
\ENSURE Optimized patch $\delta_{p_{\text{optimal}}}$

\textit{Step 1: Patch Generation}
\STATE Generate patch $\delta$ by solving:
\[
P_{\delta} = \arg\max_{\delta} \mathbb{E}_{x \sim X, t \sim T} [\log P_r(y_t | A(p, x, t))]
\]

\textit{Step 2: EigenCAM - Identify Regions}
\STATE Compute $O_{L=k} = W_{L=k}^T I$
\STATE Perform SVD: $O_{L=k} = U \Sigma V^T$
\STATE Extract $L_{\text{Eigen-Cam}} = O_{L=k} V_I$
\STATE Find key regions where $L_{\text{Eigen-Cam}}$ is high

\textit{Step 3: Grid Search - Optimize Patch Position}
\FOR{$p_i$ in candidate positions}
    \STATE Compute $C(f(I + \delta))$ at $p_i$
    \IF{$C(f(I + \delta))$ minimized}
        \STATE Update $p_{\text{optimal}}$
    \ENDIF
\ENDFOR

\textit{Step 4: Patch Optimization}
\FOR{$n = 1$ to $N$}
    \STATE Compute loss $\mathcal{L}(f(I + \delta), t) + \lambda \|\delta\|_p$
    \STATE Update $\delta \leftarrow \delta - \eta \frac{\partial \mathcal{L}}{\partial \delta}$
    \STATE Clip $\delta$ to valid pixel range
\ENDFOR

\RETURN $\delta_{p_{\text{optimal}}}$

\end{algorithmic}
\end{algorithm}

\subsection{Defense Methods}
The "Segment and Complete" (SAC) approach \cite{liu2022sac} provides an initial step in defending against adversarial patch attacks by detecting and segmenting adversarial patches. This method uses a U-Net architecture with skip connections to generate initial patch masks by identifying regions in the image that exhibit adversarial characteristics. The segmenter is trained using binary cross-entropy loss, enabling it to distinguish adversarial regions from clean areas effectively. While SAC focuses on generating accurate patch masks, these masks serve as the foundation for further processing steps, such as inpainting or diffusion-based restoration. The optimization of SAC is guided by the binary cross-entropy loss function:

\begin{equation}
\mathcal{L}_{\text{BCE}} = -\frac{1}{N} \sum_{i=1}^N \left[ y_i \log(p_i) + (1 - y_i) \log(1 - p_i) \right],
\end{equation}

where \( y_i \) represents the ground truth label, and \( p_i \) denotes the predicted probability for pixel \( i \). The training data consists of both real-world and synthetic patches. To further enhance the segmenter's robustness, a self-adversarial training procedure is employed, exposing it to increasingly challenging adversarial patches. After initial segmentation, the output masks are processed through a shape completion algorithm that refines and smoothes the detected patches, resulting in a more precise delineation of the adversarial regions.

The final masks are then applied to the input images by zeroing out the corresponding pixel values, effectively removing the adversarial patches and producing a clean, masked image for subsequent object detection. Following this, an inpainting algorithm is applied to fill in the areas identified as patches, utilizing pixel information from the surrounding regions to achieve smooth and coherent restoration. We considered two inpainting methods: the first diffuses known pixel values into the masked area while preserving texture and patterns, and the second propagates image contours (isophotes) into the masked region to maintain edge continuity. In this study, the second method was employed due to its effectiveness in seamlessly blending the removed patches into the image, thereby ensuring accurate object detection.

In addition to inpainting, we applied a stable diffusion model \cite{dhariwal2021diffusion} to further enhance image reconstruction. This generative model iteratively refines an initial noisy image into a high-quality image by reversing the diffusion process. It employs a neural network trained to predict and gradually remove noise, guiding the image toward a coherent and realistic final output. This process involves a forward diffusion step, in which noise is systematically added to the data, followed by a reverse diffusion process that methodically eliminates the noise, reconstructing the image. Specifically, the forward diffusion step is defined as:

\begin{equation}
x_t = \sqrt{\alpha_t} x_0 + \sqrt{1 - \alpha_t} \epsilon, \quad \epsilon \sim \mathcal{N}(0, I),
\end{equation}

where \( x_0 \) is the original image, \( x_t \) is the noisy image at timestep \( t \), and \( \alpha_t \) is a noise scaling factor. The reverse diffusion process is guided by a neural network \( f_\theta \) that predicts \( \epsilon \), reconstructing the image as follows:

\begin{equation}
x_{t-1} = \frac{1}{\sqrt{\alpha_t}} \left( x_t - \frac{1 - \alpha_t}{\sqrt{1 - \alpha_t}} f_\theta(x_t, t) \right) + \sqrt{1 - \alpha_t} \epsilon_t.
\end{equation}

The model operates within a latent space created by a variational autoencoder (VAE) \cite{VAE}, which allows for efficient processing during the diffusion. The stable diffusion model is constructed using a U-Net architecture with convolutional layers, skip connections, and multi-head self-attention mechanisms, which effectively capture and correlate features across different scales. Furthermore, the model incorporates cross-attention layers, conditioning the image generation on textual descriptions to achieve more precise control over the output. A key innovation of this approach is its use of a Latent Diffusion Model (LDM) that operates in a significantly reduced latent space instead of the pixel space, making the model computationally efficient while maintaining high-resolution outputs. Trained on large-scale datasets, such as LAION-5B \cite{schuhmann2022laion}, this model is capable of generating a diverse range of detailed images, offering robustness and flexibility across various image domains.

The defense methods described above are implemented in the pipeline presented in Algorithm~\ref{alg:patch_defense}, which provides a systematic approach to defend against adversarial patches by employing segmentation, inpainting, and diffusion-based methods to restore and analyze object detection performance.


\begin{algorithm}[H]
\caption{Defenses Against Adversarial Patch}
\label{alg:patch_defense}
\begin{algorithmic}[1]
\renewcommand{\algorithmicrequire}{\textbf{Input:}}  
\renewcommand{\algorithmicensure}{\textbf{Output:}}  

\REQUIRE Adversarial image $I_{\text{adv}}$, object detector $D$, SAC model, patch removal model $f_{\text{removal}}$, inpainting model $f_{\text{inpaint}}$, diffusion model $f_{\text{diffuse}}$
\ENSURE Detection results from defenses

\textit{Patch Removal Defense}
\STATE Generate binary mask $M$ using SAC (Segment and Complete) model
\STATE Apply patch removal model $f_{\text{removal}}$ using $M$ to remove the patch from $I_{\text{adv}}$
\STATE Apply object detector $D$ on restored image $I_{\text{restored}}$
\STATE Output detected class and confidence for $I_{\text{restored}}$

\textit{Patch Inpainting Defense}
\STATE Use binary mask $M$ to identify patch area in $I_{\text{adv}}$
\STATE Apply inpainting model $f_{\text{inpaint}}$ to fill in the masked area in $I_{\text{adv}}$
\STATE Apply object detector $D$ on inpainted image $I_{\text{inpainted}}$
\STATE Output detected class and confidence for $I_{\text{inpainted}}$

\textit{Patch Diffusion Defense (Stable Diffusion)}
\STATE Apply diffusion model $f_{\text{diffuse}}$ to perform denoising and remove the patch from $I_{\text{adv}}$
\STATE Perform iterative denoising steps on the patch
\STATE Apply object detector $D$ on diffused image $I_{\text{diffused}}$
\STATE Output detected class and confidence for $I_{\text{diffused}}$

\textit{Final Outputs:}
\STATE Report detection results for $I_{\text{restored}}$, $I_{\text{inpainted}}$, and $I_{\text{diffused}}$

\end{algorithmic}
\end{algorithm}

\begin{figure*}[htbp]
    \centering
    \includegraphics[width=1\textwidth]{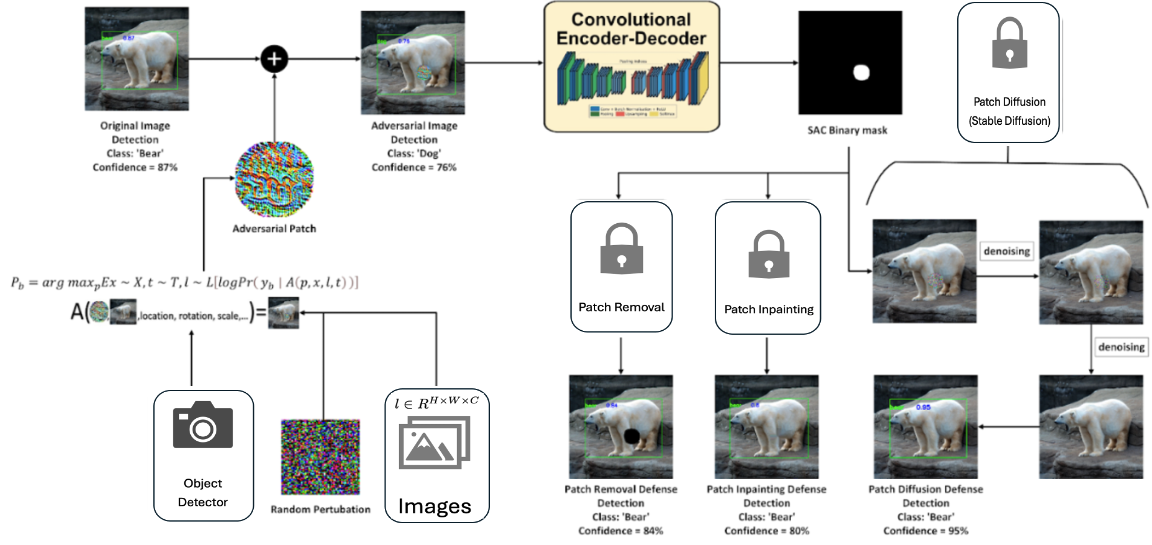}
\caption{The attacked image misclassified the bear as a dog with 76\% confidence. Defense methods corrected the prediction, restoring confidence to 84\% (patch removal), 80\% (inpainting), and 95\% (diffusion), surpassing the original confidence of 87\%.}

    \label{fig:misclassification_example}
\end{figure*}

As shown in Figure \ref{fig:misclassification_example}, the attacked image successfully deceived the model into misclassifying the bear as a dog with 76\% confidence. However, all defense methods effectively corrected the model's prediction to a bear. The patch removal defense restored the model's confidence to 84\%, while the inpainting method achieved a confidence of 80\%. The diffusion algorithm further enhanced the model's confidence to 95\%, surpassing the original confidence level of 87\%.

\subsection{Dataset}
Single-object images were chosen for this study to simplify analysis within the pipeline. Since no publicly available dataset met these requirements, a custom dataset was created by filtering the COCO dataset \cite{lin2015coco} to include only images containing one salient object. This process resulted in a dataset comprising 403 images. Additionally, a custom dataset of our own was used, which cannot be published due to the inclusion of private information. This custom dataset contains two distinct classes uniformly distributed across all images, ensuring a balanced representation for the analysis. The use of this dataset was essential for evaluating the proposed defense methods under more controlled and realistic conditions.

\section{Results}

The analysis was conducted using a systematic approach, following these steps:

\begin{enumerate}
    \item For each detection, a confidence score $C_i$ was assigned based on the predicted class. If the predicted class $\hat{y}_i$ did not match the original ground truth class $y_i$, the confidence score was set to zero:
    \begin{equation}
        C_i = 
        \begin{cases} 
            c_i, & \text{if } \hat{y}_i = y_i \\ 
            0, & \text{if } \hat{y}_i \neq y_i 
        \end{cases}
    \end{equation}
    where $c_i$ is the original confidence value output by the object detection model for the $i$-th image.
    
    \item The confidence values were then averaged across all images in each evaluation method. Let $\mathcal{C}^{(s)} = \{C_1, C_2, \dots, C_N\}$ represent the set of confidence scores for a specific method $s$, where $N$ is the total number of detections. The mean confidence $\bar{C}^{(s)}$ for method $s$ is computed as:
    \begin{equation}
        \bar{C}^{(s)} = \frac{1}{N} \sum_{i=1}^{N} C_i^{(s)}
    \end{equation}
\end{enumerate}

The average confidence values for the different methods of the pipeline are summarized in Table \ref{tab:confidence_levels_datasets}. For the original images, the mean confidence was $81.07\%$. After applying the adversarial patch, the confidence dropped to $63.18\%$. The SAC defense improved this value to $65.98\%$, while the inpainting method significantly restored it to $79.50\%$, just below the original confidence level. The diffusion defense proved to be the most effective, restoring the confidence to $84.76\%$, surpassing the original confidence. These results indicate that the adversarial patch caused a reduction in average confidence of $17.89\%$. The SAC defense yielded an improvement of $2.80\%$, while the inpainting method provided a substantial increase of $16.32\%$. The diffusion defense showed the highest efficacy, resulting in an increase of $21.58\%$ in confidence. Each change in confidence was calculated using the formula:

\begin{equation}
\frac{|Conf_{\text{average, original}} - Conf_{\text{average, i\text{th} method}}|}{Conf_{\text{average, original}}}
\end{equation}

\begin{table}[htbp]
\caption{Average Confidence Levels for Different Image Types on COCO and Custom Datasets (in percentages)}
\begin{center}
\resizebox{\columnwidth}{!}{%
\begin{tabular}{l | c c c c c}
\hline
\textbf{Dataset} & \textbf{Original} & \textbf{Patched} & \textbf{SAC} & \textbf{Inpainted} & \textbf{Diffusion} \\ \hline
COCO & 81.07\% ± 0.5 & 63.18\% ± 0.7 & 65.98\% ± 0.6 & 79.50\% ± 0.3 & \textbf{84.76\% ± 0.2} \\ 
Custom & 77.83\% ± 0.4 & 60.66\% ± 0.6 & 63.34\% ± 0.5 & 76.32\% ± 0.3 & \textbf{81.37\% ± 0.3} \\ \hline
\end{tabular}%
}
\label{tab:confidence_levels_datasets}
\end{center}
\end{table}

\begin{figure}[h]
\centering
\setlength{\tabcolsep}{1pt} 
\renewcommand{\arraystretch}{1.1} 
\begin{tabular}{ccccc}
    \textbf{Original} & \textbf{Patched} & \textbf{SAC Patched} & \textbf{Inpainted} & \textbf{Diffused} \\
    
    \includegraphics[width=0.09\textwidth]{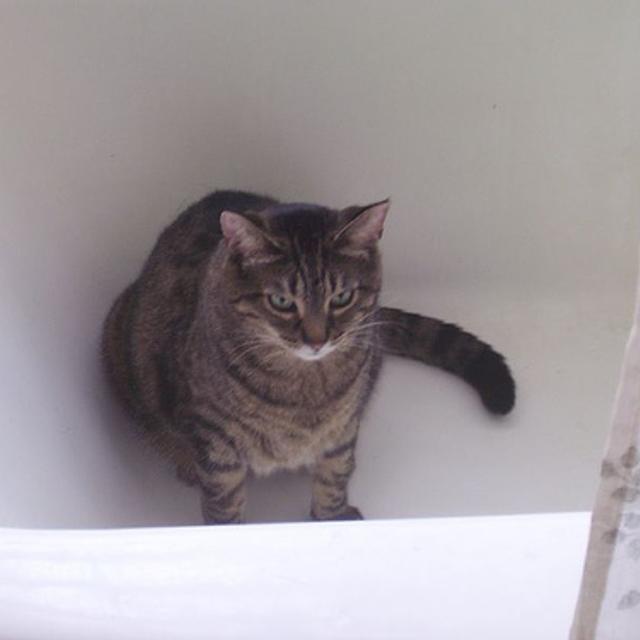} &
    \includegraphics[width=0.09\textwidth]{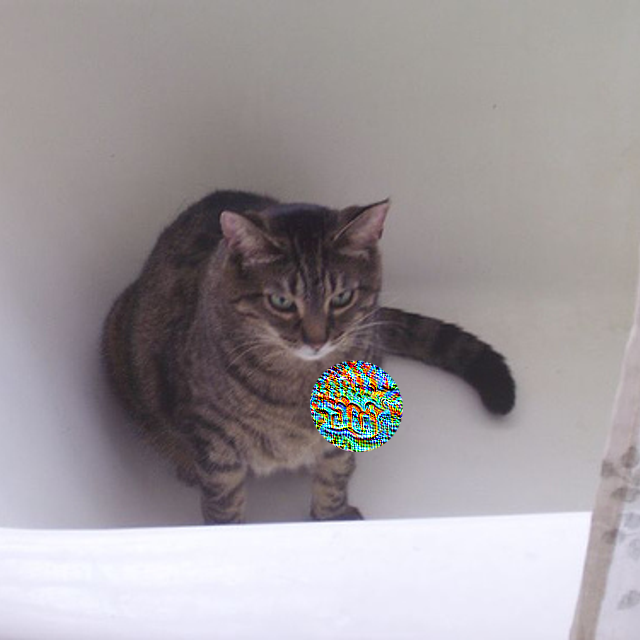} &
    \includegraphics[width=0.09\textwidth]{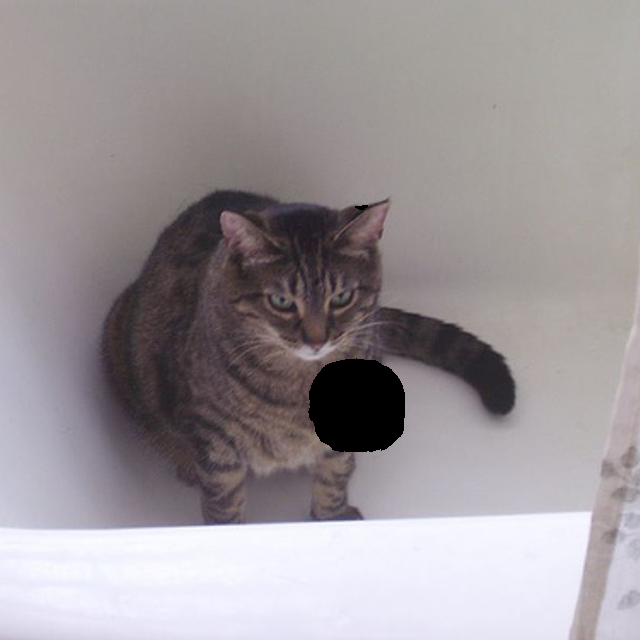} &
    \includegraphics[width=0.09\textwidth]{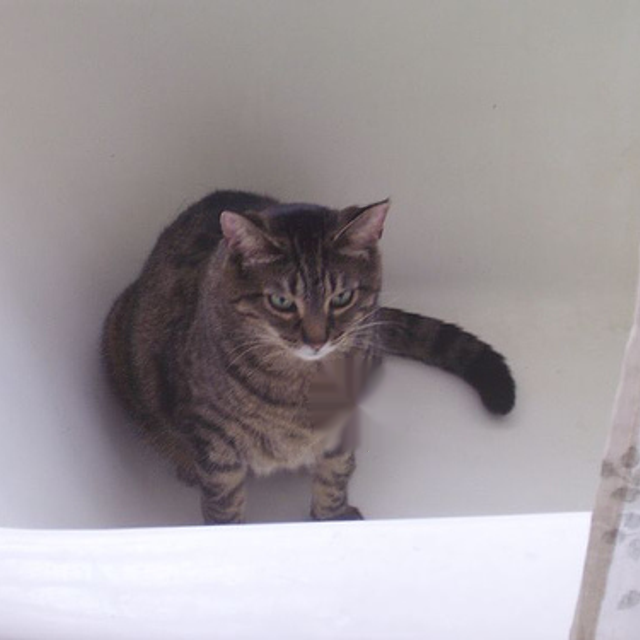} &
    \includegraphics[width=0.09\textwidth]{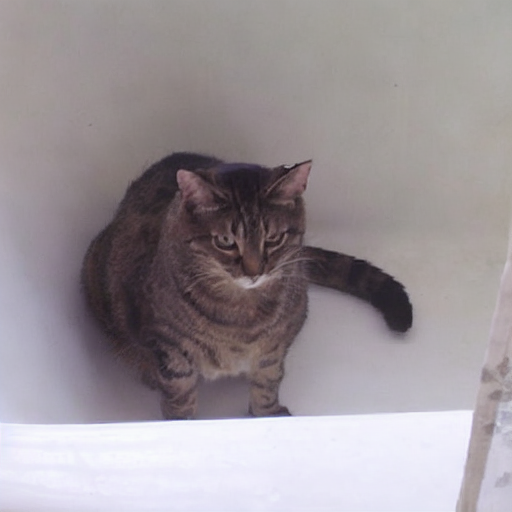} \\

    \includegraphics[width=0.09\textwidth]{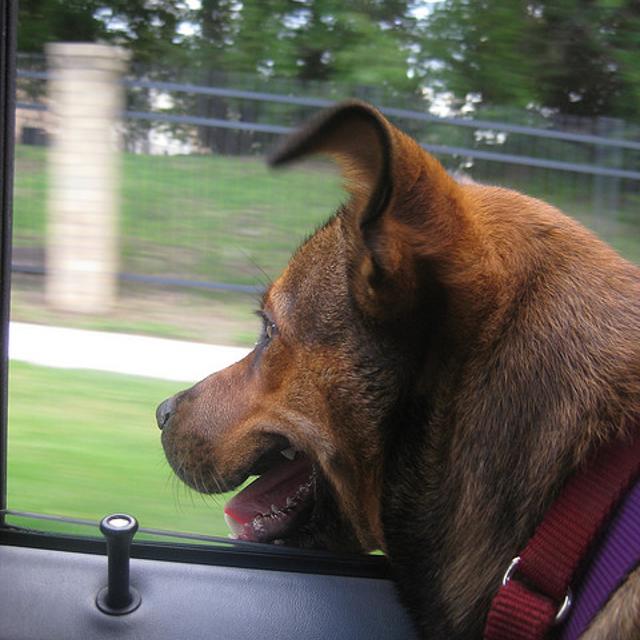} &
    \includegraphics[width=0.09\textwidth]{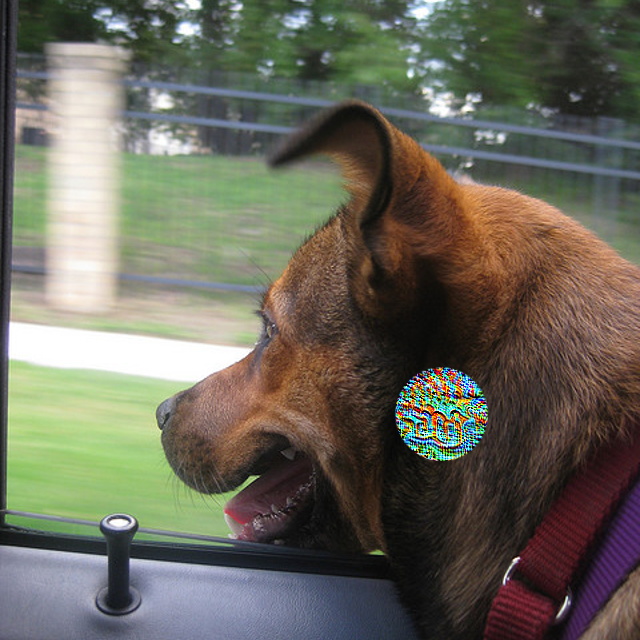} &
    \includegraphics[width=0.09\textwidth]{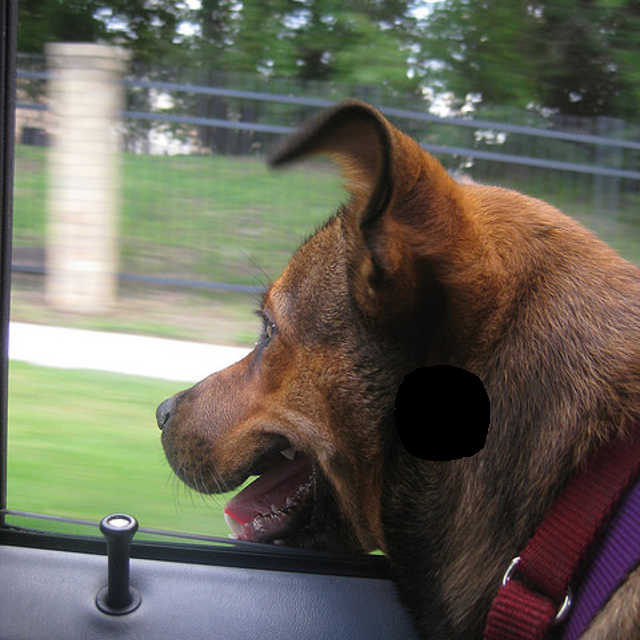} &
    \includegraphics[width=0.09\textwidth]{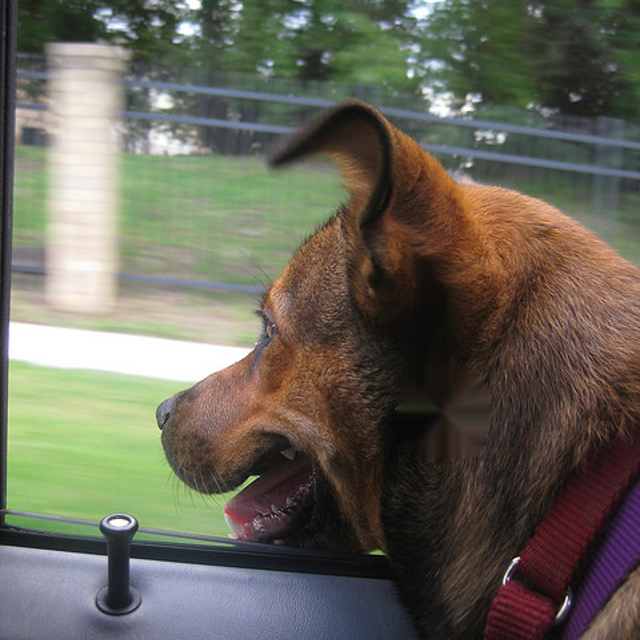} &
    \includegraphics[width=0.09\textwidth]{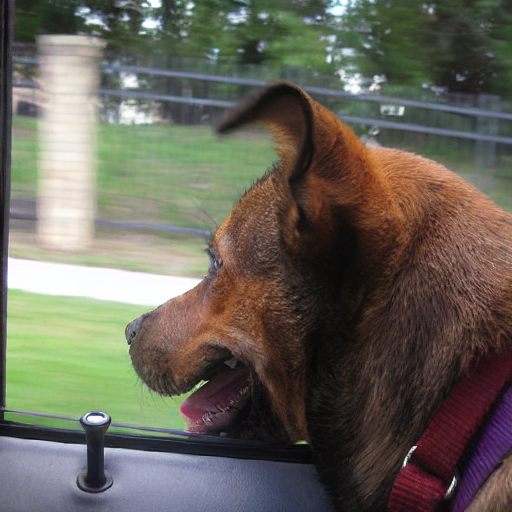} \\

    \includegraphics[width=0.09\textwidth]{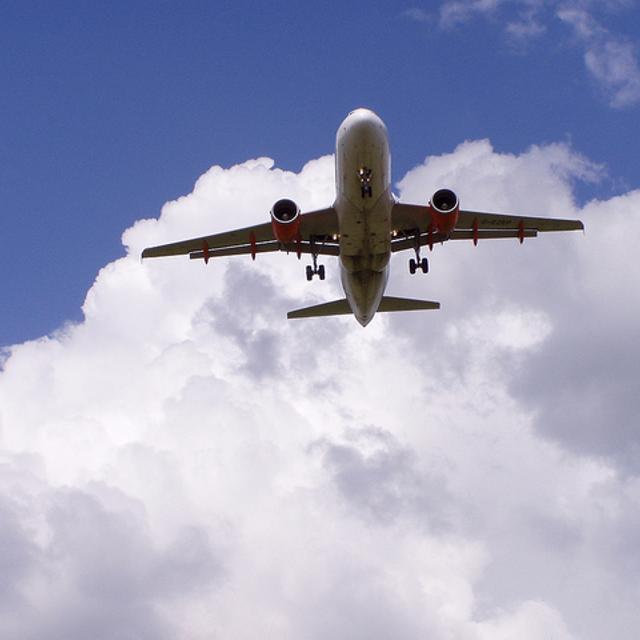} &
    \includegraphics[width=0.09\textwidth]{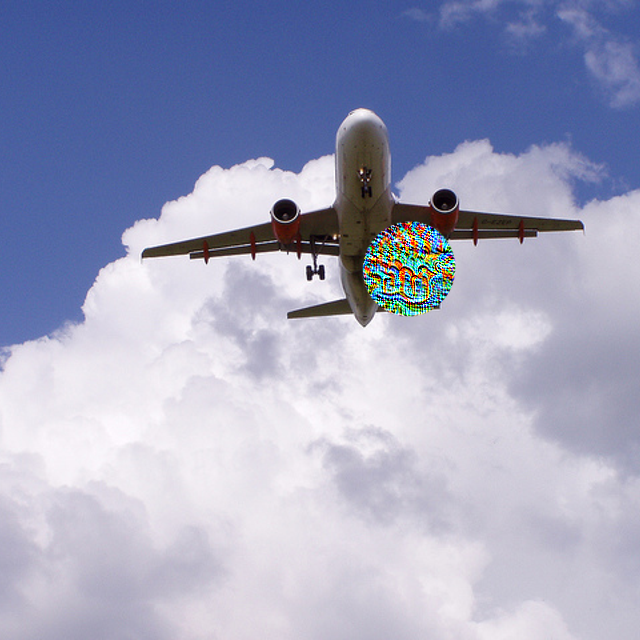} &
    \includegraphics[width=0.09\textwidth]{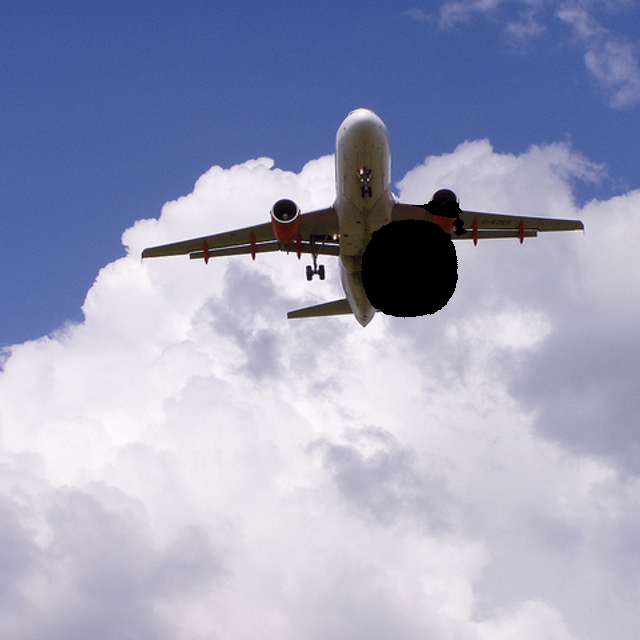} &
    \includegraphics[width=0.09\textwidth]{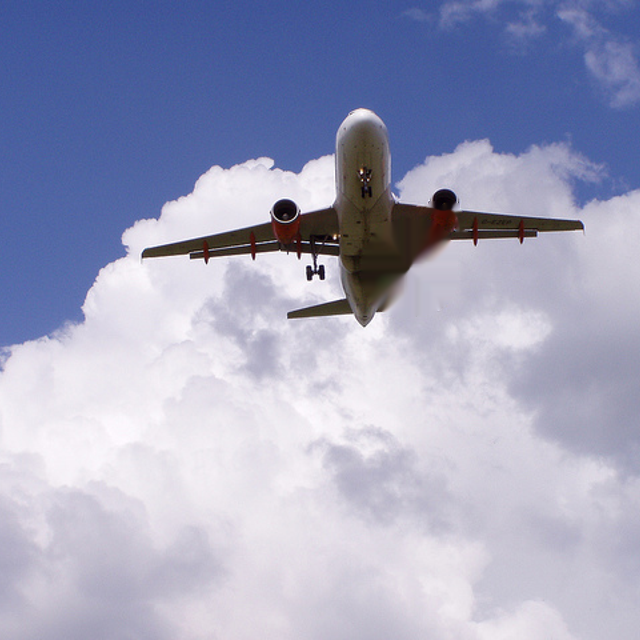} &
    \includegraphics[width=0.09\textwidth]{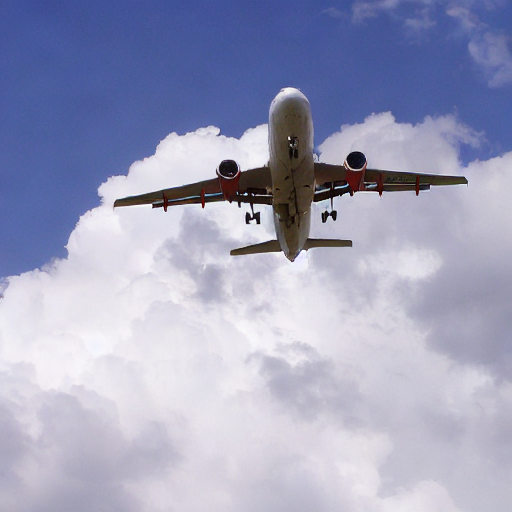} \\

    \includegraphics[width=0.09\textwidth]{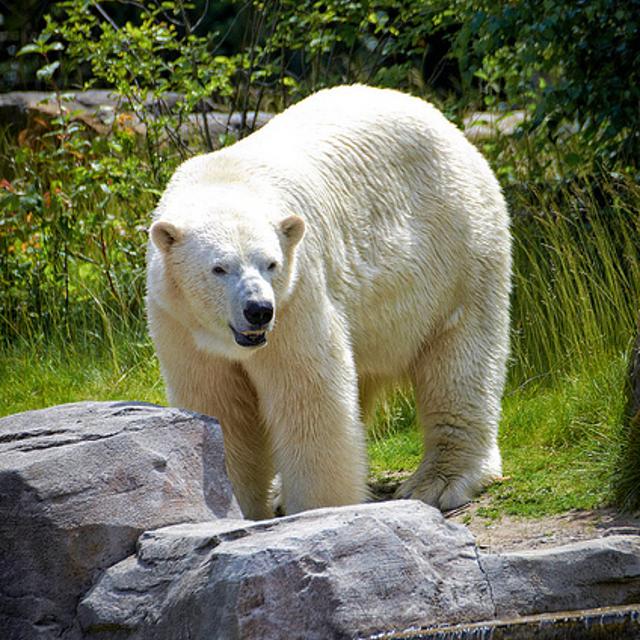} &
    \includegraphics[width=0.09\textwidth]{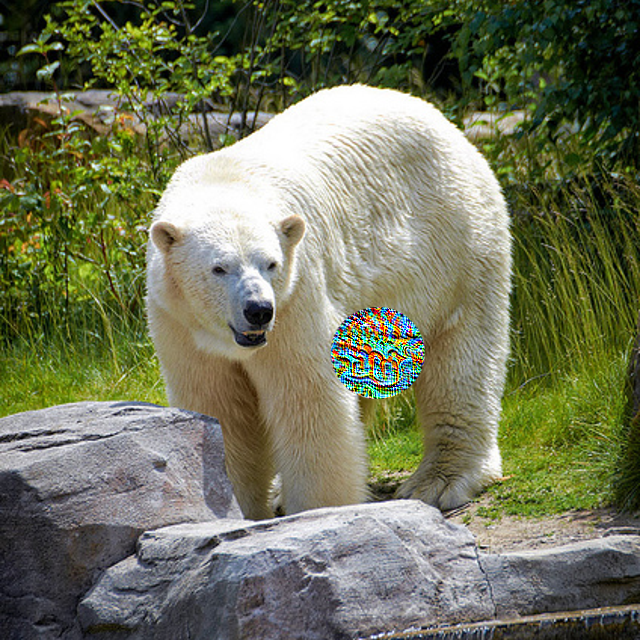} &
    \includegraphics[width=0.09\textwidth]{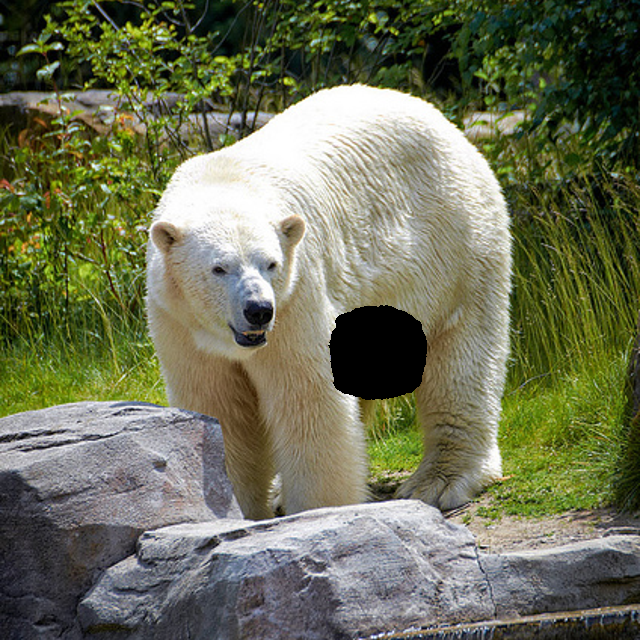} &
    \includegraphics[width=0.09\textwidth]{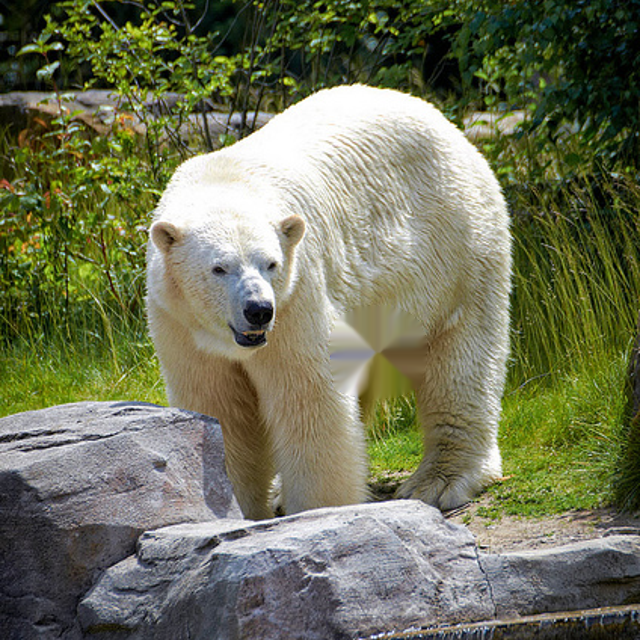} &
    \includegraphics[width=0.09\textwidth]{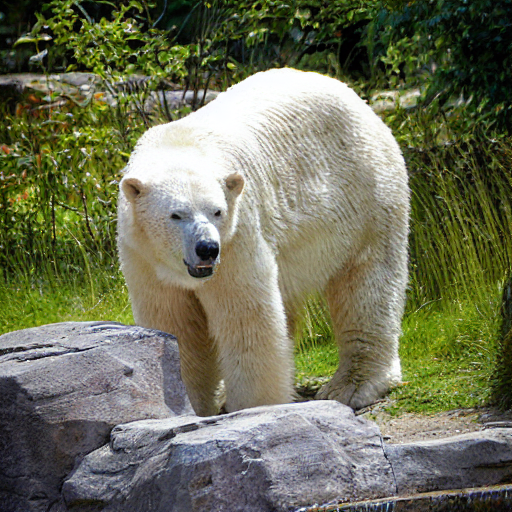} \\
\end{tabular}

\caption{Image examples at each stage: Original, Patched, SAC Patched, Inpainted, and Diffused.}
\label{fig:image_stages}
\end{figure}

Adversarial patch attacks reduce the accuracy and confidence of object detectors, making them susceptible to misclassification.
To mitigate these attacks, three key defenses are applied, as shown in Figure \ref{fig:misclassification_example}: patch removal, inpainting, and diffusion. Patch removal uses a binary mask generated by the Segment and Complete (SAC) model to locate the adversarial patch and eliminate it, restoring the image to its original state for proper object detection. Inpainting, on the other hand, uses the same binary mask but instead of removing the patch, it fills the affected area with realistic content, allowing the detector to classify objects as if the patch were never there. Finally, patch diffusion utilizes stable diffusion models to iteratively denoise the image, gradually removing the adversarial perturbation and restoring the image's integrity. This progressive refinement ensures that the image is restored, and the object detector can regain its accuracy. Algorithm \ref{alg:patch_defense} outlines the entire defense process, detailing how these methods work together to recover the detection performance after adversarial patches are applied.

\section{Ablation Study}

In this study, we conducted an ablation analysis to investigate the impact of patch shapes and sizes on the effectiveness of adversarial attacks. Initially, we experimented with multiple shapes for the adversarial patches, aiming to identify a shape that achieves a balance between attack success and practical usability. After examining various options, we selected a fixed patch shape based on its consistent performance across different scenarios. Once the shape was determined, we proceeded to investigate the effect of different patch sizes on the attack's success rate.

\begin{figure}[htbp]
    \centering
    \includegraphics[width=0.5\textwidth]{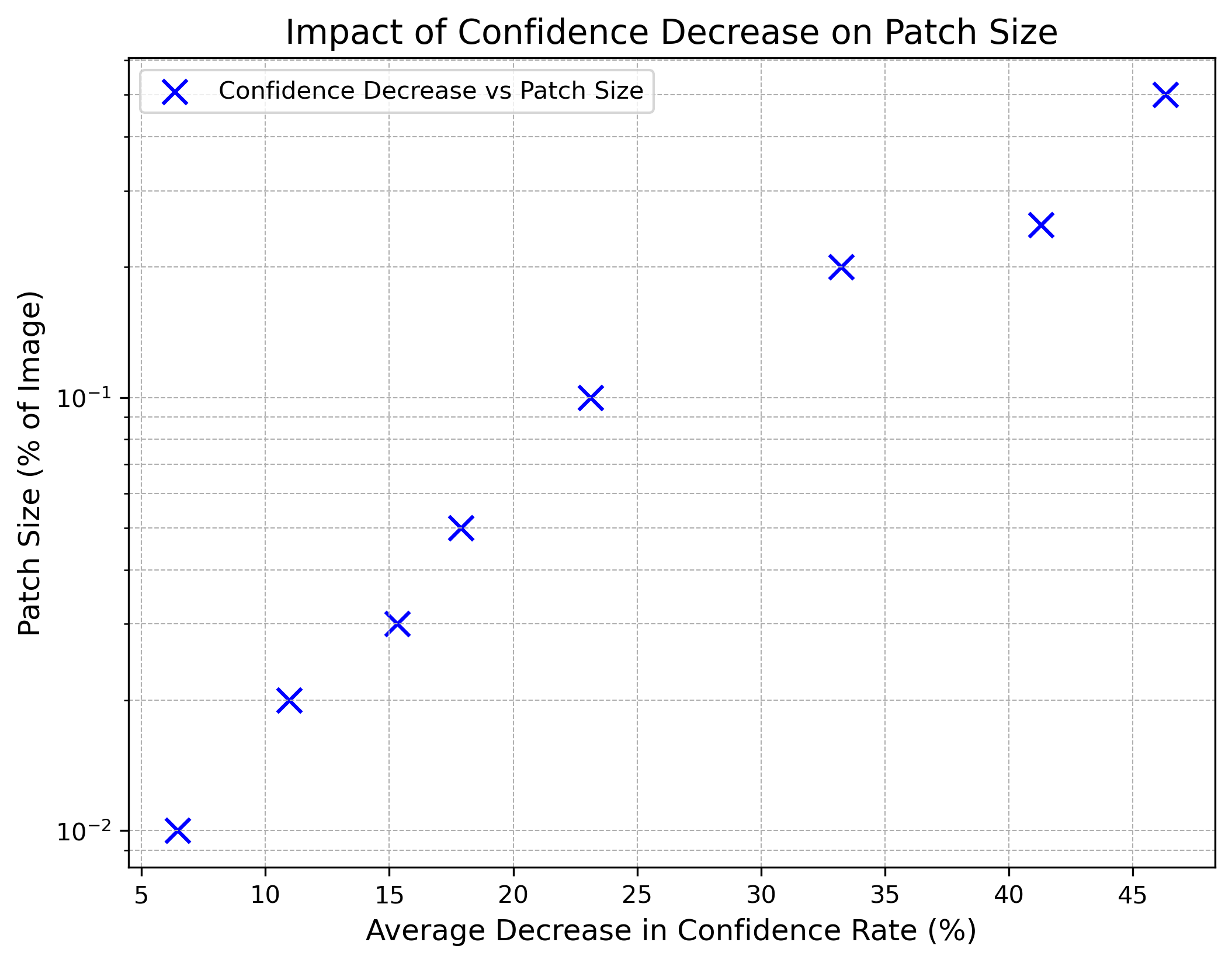}
    \caption{Impact of Confidence Decrease on Patch Size. The x-axis represents the average decrease in confidence rate as a percentage, while the y-axis shows the patch size as a percentage of the image, displayed on a logarithmic scale. Larger confidence reductions correlate with increasing patch sizes, but beyond a certain size, the marginal gains decrease.}
    \label{fig:confidence_decrease_vs_patch_size}
\end{figure}

The patch sizes were expressed as a percentage of the image, ranging from 1\% to 50\%. Our goal was to find a patch size that maximized the decrease in the object detector's confidence rate while maintaining a reasonable patch size that would not overly compromise the image's realism. Figure~\ref{fig:confidence_decrease_vs_patch_size} illustrates the results of this analysis, showing the relationship between the average decrease in confidence rate and the patch size.

As shown in Figure~\ref{fig:confidence_decrease_vs_patch_size}, larger patch sizes result in a higher decrease in confidence rates. Specifically, a patch size of 50\% caused a 46.31\% average confidence reduction, while smaller sizes such as 1\% achieved only a 6.46\% decrease. The results indicate that the effectiveness of the patch is closely tied to its size. However, beyond a patch size of approximately 20\%, the incremental improvements in attack success diminish, suggesting diminishing returns for overly large patches. Based on these observations, we identified a patch size of approximately 10\% to 20\% of the image as an optimal range for achieving a balance between attack effectiveness and practicality.

The attack generation process used in this study is built upon the framework presented in our previous work~\cite{kazoom2024improving}, where we systematically optimized the placement and configuration of the adversarial patches. Algorithm~\ref{alg:patch_attack} provides an overview of the process used to generate the attacks, including the placement strategy, optimization techniques, and evaluation metrics.

This ablation study highlights the importance of patch size in determining the effectiveness of adversarial attacks and underscores the need for careful design and evaluation to ensure both practical usability and high attack success rates.

\section{Conclusion}
The analysis indicates that the diffusion-based defense method is the most effective technique among those evaluated, as it significantly restores the confidence levels of the object detector, even surpassing the original pre-attack confidence. This finding suggests that while adversarial patches can substantially degrade the performance of object detectors, robust defense mechanisms—specifically diffusion models—can effectively counteract these attacks, enhancing detection reliability. The results underscore the potential of diffusion as a powerful tool for mitigating adversarial effects in object detection systems.

\section{Future Research}
Future research could explore several avenues to extend and generalize the current pipeline. One potential direction is to incorporate a more diverse range of adversarial patches, including those generated with varying hyperparameters designed to enhance attack efficacy, patches created using alternative neural network architectures, and naturalistic patches that closely mimic real-world textures and patterns. Additionally, investigating the impact of different patch geometries, such as squares, circles, or irregular shapes, could further refine the robustness and adaptability of the defense strategies.

Another important area for future work is testing the pipeline against more advanced object detectors, such as YOLOv7+, to evaluate the generalizability and effectiveness of the proposed defense mechanisms. Furthermore, extending the pipeline to handle images with multiple objects is a promising research direction, which could significantly broaden its applicability in complex real-world scenarios. This would involve developing and integrating new techniques for multi-object detection and defense, thereby enhancing the overall resilience of object detection models against adversarial attacks.

Lastly, while the segmentation model performs well, it is somewhat reliant on the similarity of patches to those in its training set. Exploring how to improve segmentation for patches that deviate more significantly from the training data could further enhance the robustness of the defense, though this limitation appears to have minimal impact in most cases.

\bibliographystyle{IEEEtran}
\bibliography{references}

\begin{thebibliography}{10}
\providecommand{\url}[1]{#1}
\csname url@samestyle\endcsname
\providecommand{\newblock}{\relax}
\providecommand{\bibinfo}[2]{#2}
\providecommand{\BIBentrySTDinterwordspacing}{\spaceskip=0pt\relax}
\providecommand{\BIBentryALTinterwordstretchfactor}{4}
\providecommand{\BIBentryALTinterwordspacing}{\spaceskip=\fontdimen2\font plus
\BIBentryALTinterwordstretchfactor\fontdimen3\font minus \fontdimen4\font\relax}
\providecommand{\BIBforeignlanguage}[2]{{%
\expandafter\ifx\csname l@#1\endcsname\relax
\typeout{** WARNING: IEEEtran.bst: No hyphenation pattern has been}%
\typeout{** loaded for the language `#1'. Using the pattern for}%
\typeout{** the default language instead.}%
\else
\language=\csname l@#1\endcsname
\fi
#2}}
\providecommand{\BIBdecl}{\relax}
\BIBdecl

\bibitem{hwang2023gan}
R.~H. Hwang, J.~Y. Lin, S.~Y. Hsieh, H.~Y. Lin, and C.~L. Lin, ``Adversarial patch attacks on deep-learning-based face recognition systems using generative adversarial networks,'' \emph{Sensors}, vol. 2023, 2023.

\bibitem{kazoom2025dontlag}
R.~Kazoom, R.~Lapid, M.~Sipper, and O.~Hadar, ``{Don't Lag, RAG: Training-Free Adversarial Detection Using RAG},'' \emph{arXiv preprint arXiv:2504.04858}, 2025.

\bibitem{kazoom2022meta}
\BIBentryALTinterwordspacing
R.~Kazoom, R.~Birman, and O.~Hadar, ``Meta classification model of surface appearance for small dataset using parallel processing,'' \emph{Electronics}, vol.~11, no.~21, p. 3426, 2022. [Online]. Available: \url{https://doi.org/10.3390/electronics11213426}
\BIBentrySTDinterwordspacing

\bibitem{liu2024rpa}
T.~Liu, C.~Yang, X.~Liu, R.~Han, and J.~Ma, ``Rpau: Fooling the eyes of uavs via physical adversarial patches,'' \emph{IEEE Transactions on Intelligent Transportation Systems}, vol. 2024, 2024.

\bibitem{deng2023rust}
B.~Deng, D.~Zhang, F.~Dong, J.~Zhang, M.~Shafiq, and Z.~Gu, ``Rust-style patch: A physical and naturalistic camouflage attacks on object detector for remote sensing images,'' \emph{Remote Sensing}, vol. 2023, 2023.

\bibitem{lapid2023patch}
R.~Lapid, E.~Mizrahi, and M.~Sipper, ``Patch of invisibility: Naturalistic physical black-box adversarial attacks on object detectors,'' \emph{arXiv preprint arXiv:2303.04238}, 2023.

\bibitem{lin2023diffusion}
S.~Y. Lin, E.~Chu, C.~H. Lin, J.~C. Chen, and J.~C. Wang, ``Diffusion to confusion: Naturalistic adversarial patch generation based on diffusion model for object detector,'' \emph{arXiv preprint arXiv:2307.08076}, 2023.

\bibitem{zhou2024mvp}
Z.~Zhou, H.~Zhao, J.~Liu, Q.~Zhang, G.~Wang, C.~Wang, and W.~Feng, ``Mvpatch: More vivid patch for adversarial camouflaged attacks on object detectors in the physical world,'' \emph{arXiv preprint arXiv:2312.17431}, 2024.

\bibitem{wei2022survey}
H.~Wei, H.~Tang, X.~Jia, Z.~Wang, H.~Yu, Z.~Li, S.~Satoh, L.~Van~Gool, and Z.~Wang, ``Physical adversarial attack meets computer vision: A decade survey,'' \emph{arXiv preprint arXiv:2209.15179}, 2022.

\bibitem{chua2022duet}
T.~J. Chua, W.~Yu, C.~Liu, and J.~Zhao, ``Detection of uncertainty in exceedance of threshold (duet): An adversarial patch localizer,'' in \emph{IEEE/ACM International Conference on Big Data Computing, Applications and Technologies (BDCAT)}, 2022.

\bibitem{liu2022sac}
J.~Liu, A.~Levine, C.~H.~L. Lau, R.~Chellappa, and S.~Feizi, ``Segment and complete: Defending object detectors against adversarial patch attacks with robust patch detection,'' in \emph{Proceedings of the IEEE/CVF Conference on Computer Vision and Pattern Recognition (CVPR)}, 2022, pp. 14\,973--14\,982.

\bibitem{kim2022ape}
T.~Kim, Y.~Yu, and Y.~M. Ro, ``Defending physical adversarial attack on object detection via adversarial patch-feature energy,'' in \emph{Proceedings of the 30th ACM International Conference on Multimedia}, 2022.

\bibitem{yang2023ibcd}
D.~Yang, Y.~Huang, Q.~Guo, F.~Juefei-Xu, M.~Hu, Y.~Liu, and G.~Pu, ``Architecture-agnostic iterative black-box certified defense against adversarial patches,'' \emph{arXiv preprint arXiv:2305.10929}, 2023.

\bibitem{kang2024diffender}
C.~Kang, Y.~Dong, Z.~Wang, S.~Ruan, H.~Su, and X.~Wei, ``Diffender: Diffusion-based adversarial defense against patch attacks,'' \emph{arXiv preprint arXiv:2306.09124}, 2024.

\bibitem{xue2024dpg}
Y.~Xue, M.~Wen, W.~He, and W.~Li, ``Dpg: A model to build feature subspace against adversarial patch attack,'' \emph{Machine Learning}, vol. 2024, pp. 1--22, 2024.

\bibitem{liang2024texture}
J.~Liang, R.~Yi, J.~Chen, Y.~Nie, and H.~Zhang, ``Securing autonomous vehicles visual perception: Adversarial patch attack and defense schemes with experimental validations,'' \emph{IEEE Transactions on Intelligent Vehicles}, vol. 2024, 2024.

\bibitem{pathak2024agnostic}
S.~Pathak, S.~Shrestha, and A.~AlMahmoud, ``Model agnostic defense against adversarial patch attacks on object detection in unmanned aerial vehicles,'' \emph{arXiv preprint arXiv:2405.19179}, 2024.

\bibitem{lin2024nutnet}
Z.~Lin, Y.~Zhao, K.~Chen, and J.~He, ``I don't know you, but i can catch you: Real-time defense against diverse adversarial patches for object detectors,'' \emph{arXiv preprint arXiv:2406.10285}, 2024.

\bibitem{kazoom2024improving}
R.~Kazoom, R.~Birman, and O.~Hadar, ``Improving the robustness of object detection and classification ai models against adversarial patch attacks,'' \emph{arXiv preprint arXiv:2403.12988}, 2024.

\bibitem{muhammad2020eigen}
M.~B. Muhammad and M.~Yeasin, ``Eigen-cam: Class activation map using principal components,'' in \emph{2020 International Joint Conference on Neural Networks (IJCNN)}.\hskip 1em plus 0.5em minus 0.4em\relax IEEE, 2020.

\bibitem{dhariwal2021diffusion}
P.~Dhariwal and A.~Nichol, ``Diffusion models beat gans on image synthesis,'' in \emph{Advances in Neural Information Processing Systems}, vol.~34, 2021, pp. 8780--8794.

\bibitem{VAE}
C.~Doersch, ``Tutorial on variational autoencoders,'' \emph{arXiv preprint arXiv:1606.05908}, 2021.

\bibitem{schuhmann2022laion}
C.~Schuhmann, R.~Beaumont, R.~Vencu, C.~Gordon, R.~Wightman, A.~Valluri, J.~Jin, Y.~Kirillov, A.~Shtern, A.~Katta \emph{et~al.}, ``Laion-5b: An open large-scale dataset for training next generation image-text models,'' in \emph{Advances in Neural Information Processing Systems}, vol.~35, 2022, pp. 25\,278--25\,294.

\bibitem{lin2015coco}
T.~Y. Lin, M.~Maire, S.~Belongie, J.~Hays, P.~Perona, D.~Ramanan, C.~L. Zitnick, and P.~Dollar, ``Microsoft coco: Common objects in context,'' in \emph{Proceedings of the IEEE Conference on Computer Vision and Pattern Recognition (CVPR)}, 2015, pp. 740--755.

\end{thebibliography}

\end{document}